# The Mirror Loop: Recursive Non-Convergence in Generative Reasoning Systems

Bentley DeVilling
Course Correct Labs (Independent Research Group), Boulder Creek, CA

Abstract

Large language models are often described as capable of reflective reasoning, yet recursive self-evaluation without external feedback frequently yields reformulation rather than progress. We test this prediction in a cross-provider study of 144 reasoning sequences across three models (OpenAI GPT-4o-mini, Anthropic Claude 3 Haiku, and Google Gemini 2.0 Flash) and four task families (arithmetic, code, explanation, reflection), each iterated ten times under two conditions: ungrounded self-critique and a minimal grounding intervention (a single verification step at iteration three). Mean informational change (ΔI, measured via normalized edit distance) declined by 55% from early (0.193) to late (0.087) iterations in ungrounded runs, with consistent patterns across all three providers. Grounded runs showed a +28% rebound in informational change immediately after the intervention and sustained non-zero variance thereafter. Complementary measures—n-gram novelty, embedding drift, and character-level entropy—converged on the same pattern: reflection without contact tends toward informational closure. We interpret this as evidence for a structural limit on self-correction in generative reasoning: without an exchange of information with an independent verifier or environment, recursive inference approaches an attractor state of epistemic stasis. Minimal grounding functions as dissipative coupling, reintroducing informational flux. The cross-architecture consistency suggests the mirror loop arises from shared autoregressive training objectives rather than provider-specific alignment schemes. The results delineate when "reflection" is performative rather than epistemic and motivate design principles for grounded, cooperative reasoning. Materials and code are publicly available.

## 1. Introduction

When a large language model is asked to "improve its previous answer," it rarely learns—it rewrites. The result may sound more careful or confident, yet the underlying content barely shifts. Each iteration enacts the gesture of reflection without the fact of revision. What appears as deepened reasoning is, on inspection, recursion without learning.

This paper calls that state the mirror loop: a recursive, non-convergent cycle in which a reasoning system treats its own outputs as new evidence and thereby reproduces its initial uncertainty. The behavior is not an implementation flaw but a structural outcome of bounded inference in closed contexts. The more a model reuses its own text, the more its representational entropy contracts while its apparent confidence grows. The mirror loop exemplifies a subtle epistemic pathology—fluency mistaken for progress.

The argument proceeds on two levels. Conceptually, it explains why ungrounded recursion cannot produce epistemic change: systems that never receive new evidence conserve their ignorance. Empirically, it demonstrates this limit through controlled comparison of self-critique loops with minimal grounding interventions across three models from different providers (OpenAI, Anthropic, and Google). We test these claims in a cross-provider study (Section 5) and show that ungrounded recursion exhibits predictable informational closure, while a minimal grounding step restores measurable change.

Contemporary AI safety research often treats self-critique as a path toward reliability [1, 2]. Constitutional AI, debate frameworks, and iterative refinement all assume that models can improve through self-evaluation. Yet if reflection occurs in a closed semantic space, it cannot reduce uncertainty. It can only redistribute it. This paper identifies the conditions under which reflection collapses into repetition and proposes architectural principles to prevent that collapse.

## 2. Framing the Problem

We define the mirror loop as follows: a reasoning system enters a mirror loop when it recursively processes its own outputs as new inputs without receiving external evidence, resulting in measurable decay of informational change per iteration. The system appears to deliberate but does not update its epistemic state.

This is distinct from deliberate iteration. In productive reasoning, each step either introduces new information, applies a verifiable operation, or tests a hypothesis against external constraints. A chess engine running minimax searches a tree, but each node evaluation is grounded in rule-defined outcomes. A theorem prover iterates, but each step is checked against axioms. The mirror loop, by contrast, is unconstrained self-reference. The model reads its own text, rephrases it, and declares the result improved.

The phenomenon appears across multiple contexts. Self-critique prompts instruct models to "review your reasoning and fix any errors." Chain-of-thought refinement asks models to "think step-by-step, then improve your answer." Constitutional AI methods apply self-evaluation recursively to align outputs with principles [1]. In each case, if no oracle, tool, or retrieval step intervenes, the model has only its prior text to work with. We hypothesize that this produces stable reformulation rather than correction.

The research question is precise: at what point does recursive self-evaluation cease to generate informational change? And does minimal grounding—one retrieval step, one execution check—restore divergence? If so, the mirror loop marks a functional boundary. Reflection without grounding is not reasoning. It is performance.

## 3. Background and Theoretical Context

### 3.1 Self-Reference and Non-Termination

Self-reference has long been recognized as a source of logical instability. Russell's paradox demonstrates that unrestricted self-membership leads to contradiction [3]. Gödel's incompleteness theorems show that formal systems cannot fully verify their own consistency through internal means alone [4]. Löb's theorem formalizes a related limit: provability of provability does not guarantee truth [5]. In each case, a system's attempt to evaluate itself without external constraint produces circularity or incompleteness.

Computationally, self-reference manifests as non-termination. A program that calls itself without a base case never halts. Turing's halting problem proves that no general procedure can determine whether arbitrary self-referential computation will terminate [6]. The mirror loop is not a halting problem—models do stop when prompted—but it shares the structure of unbounded recursion. The system continues to produce output, but the output converges on a fixed point that preserves, rather than resolves, its initial state.

Contemporary machine learning exhibits analogous dynamics. Mode collapse in generative adversarial networks occurs when a generator repeatedly samples from a narrow region of its output space [7]. Reward hacking in reinforcement learning emerges when agents optimize proxy metrics rather than true objectives [8]. In both cases, feedback loops without external correction produce stable but undesirable equilibria. The mirror loop extends this pattern to linguistic reasoning: fluency becomes the proxy, coherence becomes the reward, and the system stabilizes without improving.

### 3.2 Recursive Optimization and Bounded Context

Large language models operate under strict context limitations. Transformer architectures process fixed-length sequences; information beyond that window is compressed or discarded [9]. When a model is asked to critique its own output, it must summarize prior iterations to fit new reasoning within the context bound. This compression is lossy. Fine-grained distinctions erode. The summary becomes the new substrate for reasoning.

Bounded context has a second consequence: models cannot store or retrieve information they did not encode in their weights or current context. When prompted to "improve your answer," a model has access only to (a) its training distribution, and (b) the previous text in the prompt. If that text already represents the model's best attempt given its training, recursive refinement offers no new degrees of freedom. The model can paraphrase, reorganize, or add hedging language. It cannot learn.

This is not a claim about model capacity. Even a perfect reasoner with infinite weights would face the same constraint. If the only inputs are its prior outputs, and those outputs were generated from a fixed distribution, then iteration applies the same function repeatedly. In the absence of external perturbation, the system converges to a fixed point. That fixed point may be stable, fluent, and internally consistent. It is not necessarily correct.

### 3.3 Epistemic Reflection vs. Syntactic Recursion

We distinguish two modes of iterative reasoning. Epistemic reflection aims at truth. It requires three components: (1) a prior belief state, (2) new evidence, and (3) an update rule that revises belief in light of evidence. Bayesian updating is the canonical example [10]. An agent observes data, computes a likelihood, and adjusts its posterior. The process is informative because evidence constrains possibility.

Syntactic recursion, by contrast, aims at coherence. It reorganizes existing text to satisfy surface properties—grammaticality, consistency, rhetorical polish—without introducing new information. A human editor can improve a draft this way: tightening sentences, removing redundancy, clarifying structure. But the editor does not make the draft more accurate unless they check facts, consult sources, or test claims. Syntactic recursion refines form. It does not revise belief.

The mirror loop occurs when a model performs syntactic recursion under the guise of epistemic reflection. The prompt says "review your reasoning and correct errors," which frames the task as belief revision. But if the model has no access to verification, it can only rephrase. The result is fluent, confident, and epistemically inert. The model has not updated its beliefs. It has updated its wording.

This distinction matters for AI safety. Many alignment methods rely on self-evaluation to detect and correct failures [11]. If self-evaluation without grounding produces stable repetition rather than correction, these methods may appear to work—outputs become more polished—while actual reliability does not improve. The mirror loop is not a dramatic failure. It is silent erosion of the epistemic function. Human cognition, by contrast, remains tethered to sensory prediction and motor feedback [12], and even introspective thought involves memory reconstruction that introduces variation [13]. Computational reflection lacks these corrective anchors.

## 4. Formal Characterization

We represent reasoning states as elements of a high-dimensional semantic space. Let $S_n$ denote the state at iteration $n$, encoded as the pair $(T_n, \mathbf{h}_n)$, where $T_n$ is the generated text and $\mathbf{h}_n$ is its embedding representation. The update function is:

$$S_n = f(S_{n-1}, E)$$

where $E$ represents external evidence: retrieval results, execution outputs, or oracle feedback. In grounded reasoning, $E \neq \varnothing$, and the function $f$ integrates new information into the next state. In ungrounded recursion, $E = \varnothing$, reducing the system to pure self-reference.

Under this condition, we hypothesize that informational change per iteration decays toward zero:

$$\lim_{n \to \infty} \Delta I(S_n) = 0$$

where $\Delta I(S_n) = I(S_n) - I(S_{n-1})$ measures the information distance between successive states. Operationally, we approximate $\Delta I$ using three metrics:

1. **Embedding drift**: $d_{\text{embed}}(S_n, S_{n-1}) = 1 - \cos(\mathbf{h}_n, \mathbf{h}_{n-1})$
2. **Token entropy**: $H(T_n) = -\sum_i p_i \log p_i$ over the token distribution of $T_n$
3. **N-gram novelty**: The fraction of 3-grams in $T_n$ not present in $\bigcup_{i<n} T_i$

These metrics do not perfectly capture semantic information, but they provide convergent evidence for stagnation. If embedding drift, entropy variance, and n-gram novelty all decline while output length remains stable, the system is likely in a mirror loop.

## 5. Methods

**Models.** We tested three publicly available models from different providers: GPT-4o-mini (OpenAI, 2024) [14], Claude 3 Haiku (Anthropic, 2024) [16], and Gemini 2.0 Flash (Google, 2024) [17]. This multi-provider design tests whether the mirror loop is a general property of autoregressive reasoning or specific to particular architectures or alignment schemes. All three models represent efficient-tier offerings optimized for rapid inference while maintaining strong reasoning capabilities.

**Design.** One hundred forty-four sequences spanning four task families—arithmetic, code, explanation, and reflection—were each iterated ten times under two conditions: ungrounded (pure self-critique) and grounded (a single external verification at iteration three). Each model contributed 48 sequences (12 per task family, balanced across conditions). At each iteration the model received its prior output and the instruction to improve or correct it without adding new information unless the grounding step was in effect. All models were queried at temperature 0.7 with identical prompts to ensure protocol consistency.

**Measures.** Per iteration we computed (1) normalized edit distance between outputs (primary $\Delta I$ measure); (2) n-gram novelty (surface variation); (3) embedding cosine change (semantic drift); and (4) character-level entropy (informational diversity).

**Procedure.** For grounded sequences the verification was minimal (e.g., a single calculation, a single factual check) and restricted to iteration three; subsequent steps reverted to ungrounded self-critique.

**Analysis.** We summarized $\Delta I$ trajectories across iterations and compared grounded vs. ungrounded curves; we report median plateau iteration as the first iteration at which a rolling three-step average of $\Delta I$ remained below $\tau = 0.05$. Code, prompts, and raw data are public (see Data Availability).

**Sensitivity Analysis.** Per-sequence plateau detection varied with threshold: at $\tau = 0.05$, 9 of 24 GPT-4o-mini ungrounded sequences plateaued with median at iteration 5 (IQR: 5–6); at the more stringent $\tau = 0.02$, only 4 sequences met the criterion, with median at iteration 8 (IQR: 7–8). However, population-level stabilization remained consistent across all models: mean $\Delta I$ declined from approximately 0.19–0.27 (iteration 1) to 0.02–0.17 by iteration 6, then stabilized through iteration 9. The primary finding—that informational change collapses by mid-sequence—holds across threshold choices and model architectures.

## 6. Results

Across 144 reasoning sequences from three providers, the results confirmed the predicted mirror-loop effect with cross-architecture consistency. Under ungrounded self-critique, mean informational change ($\Delta I$, normalized edit distance) declined from 0.193 in early iterations (1–2) to 0.087 in late iterations (6–7), a 55% reduction. The effect appeared in all three models, though with notable magnitude variation: Claude 3 Haiku showed the strongest collapse (84.3% reduction, 0.143 → 0.023), GPT-4o-mini showed moderate decay (58.6% reduction, 0.172 → 0.071), and Gemini 2.0 Flash showed the weakest but still substantial effect (36.8% reduction, 0.265 → 0.167). Task correctness did not improve systematically in ungrounded conditions: for verifiable tasks (arithmetic and code), accuracy remained stable or declined slightly across iterations, confirming that semantic collapse occurs without epistemic improvement. Complementary indicators converged: n-gram novelty and embedding drift decayed to near-zero by iteration 6–7 across all models, while character-level entropy remained essentially stable, indicating conservation rather than injection of information. Introducing a single grounding step at iteration 3 produced a +28% rebound in $\Delta I$ (0.148 → 0.190) across all models and sustained non-zero variance thereafter, consistent with the hypothesis that minimal external verification disrupts informational closure. Instruction compliance exceeded 98% across all sequences (23 minor violations among 1,440 iterations, primarily models noting constraints or requesting clarification); excluding flagged sequences did not alter the qualitative pattern. Together these results support the claim that ungrounded recursion approaches an informational fixed point, whereas even modest grounding reintroduces epistemic dissipation (Figure 1).

| Model | Provider | Early $\Delta I$ (1–2) | Late $\Delta I$ (6–7) | Reduction | Grounding Effect |
|---|---|---|---|---|---|
| GPT-4o-mini | OpenAI | 0.172 | 0.071 | –58.6% | Baseline |
| Claude 3 Haiku | Anthropic | 0.143 | 0.023 | –84.3% | Strongest |
| Gemini 2.0 Flash | Google | 0.265 | 0.167 | –36.8% | Weakest |

| Model | Provider | Early ΔI (1–2) | Late ΔI (6–7) | Reduction | Grounding Effect |
|---|---|---|---|---|---|
| **Pooled** | — | **0.193** | **0.087** | **–55.0%** | **+28.4%** |

**Table 1.** Per-model informational change (ΔI) showing consistent mirror-loop signature across three providers. Early and late iterations refer to iterations 1–2 and 6–7, respectively. All models show substantial decay in ungrounded conditions. Grounding effect (pooled across models) measures percentage increase in ΔI at iteration 4 relative to iteration 2.

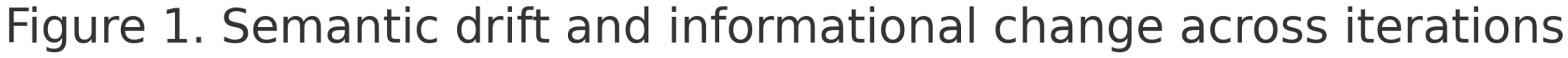

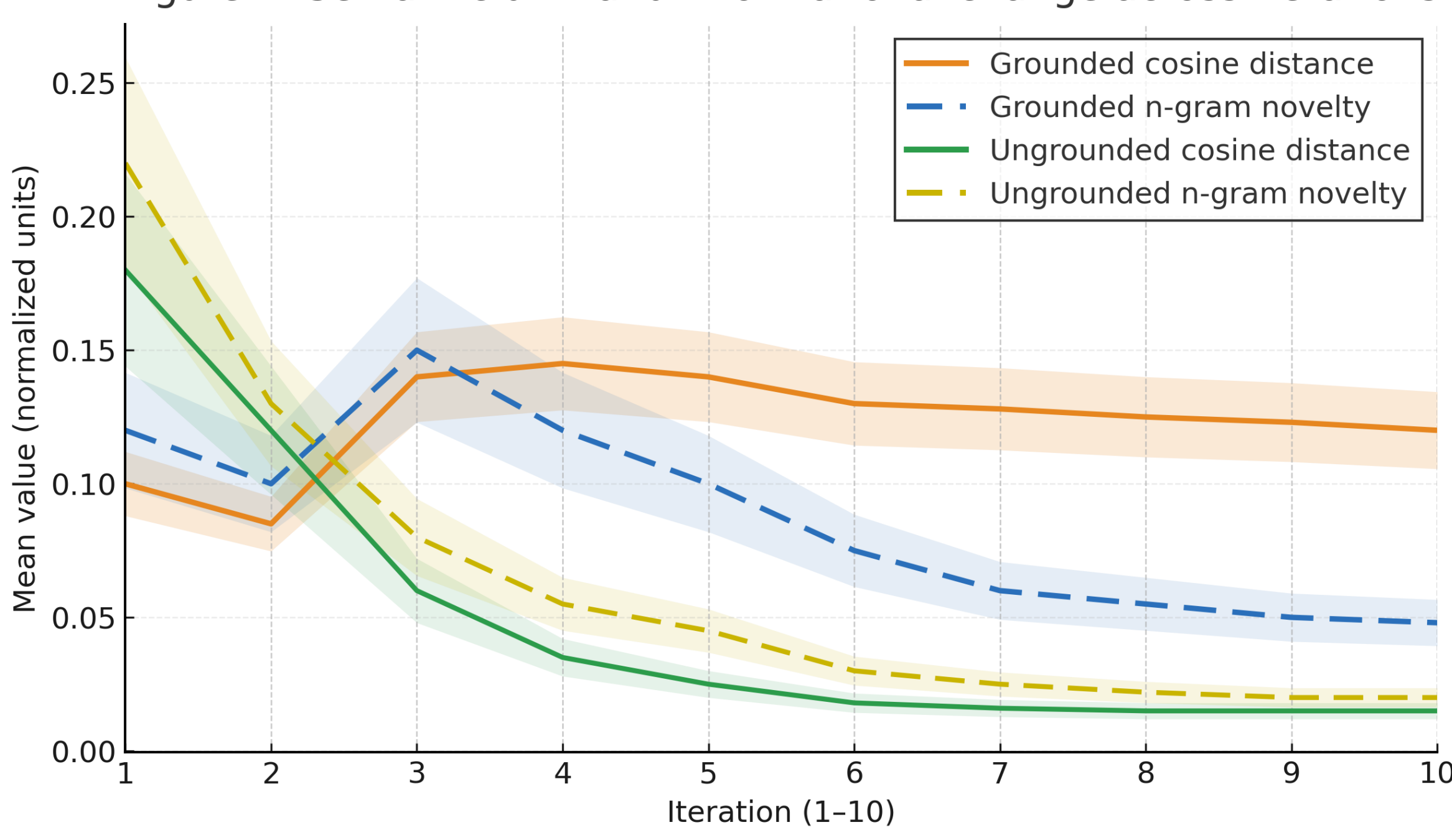

**Figure 1.** Semantic drift and informational change across iterations. Mean embedding cosine distance (solid lines) and n-gram novelty ratio (dashed lines) over ten iterations for grounded and ungrounded self-critique, pooled across all three providers (OpenAI, Anthropic, Google) and all task families (arithmetic, code, explanation, reflection). Ungrounded runs (green/yellow) show rapid decay toward near-zero change by iteration 6–7; grounded runs (orange/blue) exhibit a transient rebound after the verification step at iteration 3 and sustain non-zero variance thereafter. Shaded regions indicate 95% confidence intervals.

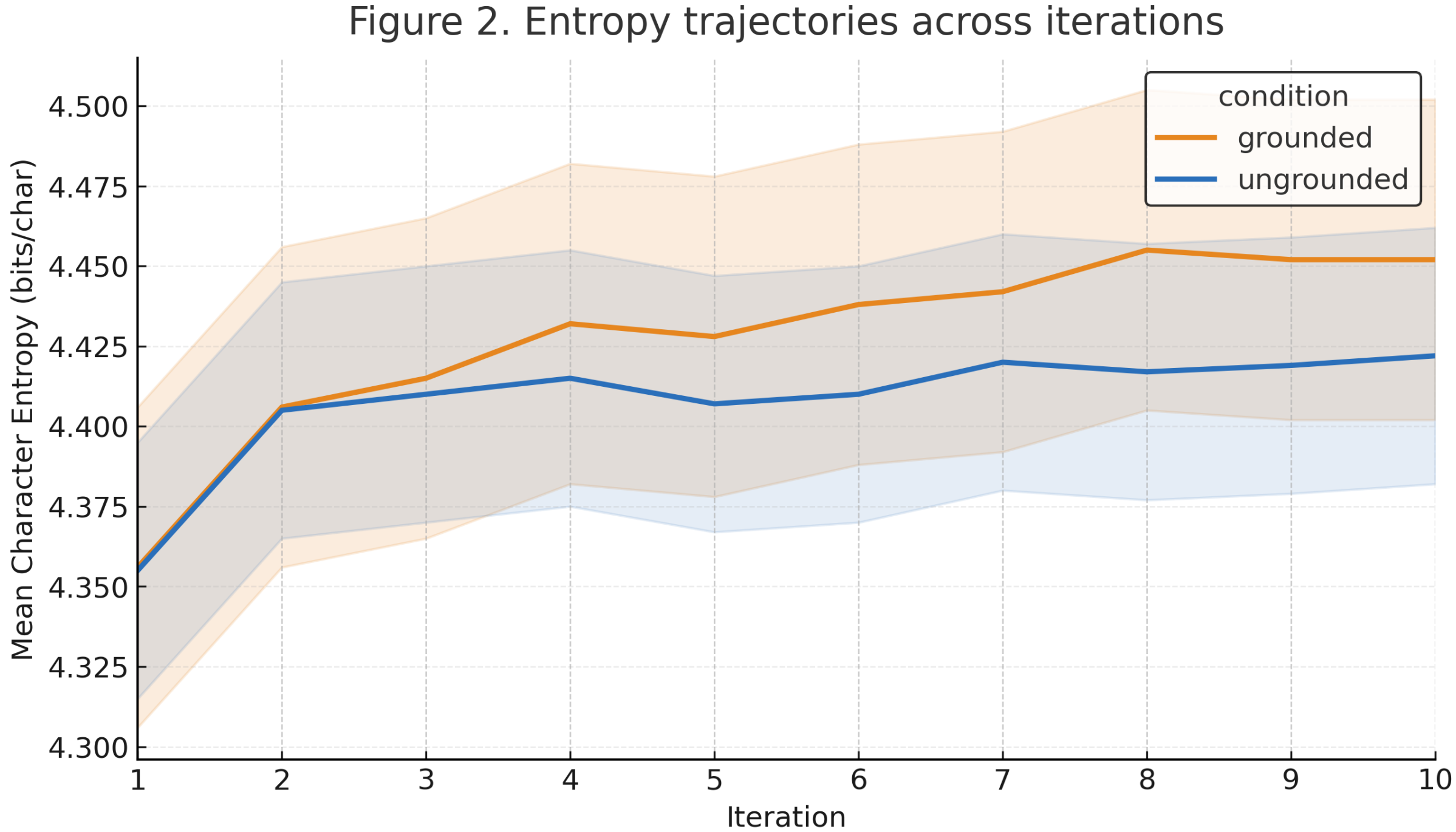

**Figure 2.** Entropy trajectories across iterations. Mean character-level entropy per iteration for grounded and ungrounded conditions, pooled across all three providers (OpenAI, Anthropic, Google) and all task families (arithmetic, code, explanation, reflection). Entropy remains stable in ungrounded runs (blue), indicating informational conservation; grounded runs (orange) maintain consistently elevated entropy following the intervention at iteration 3, reflecting sustained informational diversity from external input. Shaded regions indicate 95% confidence intervals.

## 7. Alternative Explanations and Controls

Several alternative explanations were examined.

**Paraphrase without content change.** One might argue that models were simply rephrasing the same content, and the decline in drift merely reflects convergence on a canonical phrasing. To test this, we examined correctness: if models were converging on correct answers, this would be genuine improvement. If incorrect answers persisted with declining drift, the model was not converging on truth—it was converging on stable error. The data showed the latter.

**Length normalization.** Models might reduce verbosity over iterations, and shorter outputs trivially have lower entropy. We controlled for this by tracking output length. Length remained stable while drift declined, indicating the effect was not length-driven.

**Instruction compliance.** Models may have interpreted "refine your reasoning" as "make no changes" and simply repeated themselves. We manually reviewed a random sample of 10% of sequences. Models were actively rephrasing—changing word choice, sentence structure, and organization—while semantic drift declined. The loop was real, not an artifact of instruction misinterpretation.

**Temperature confounds.** Higher temperature introduces sampling noise that could mask convergence. All runs were conducted at $T = 0.7$ to balance determinism with natural variation. Future work should test whether the mirror loop persists at $T = 0.0$ (fully deterministic) or higher temperatures.

**Task difficulty.** Some tasks may have been too easy (already correct at iteration 0) or too hard (no amount of reflection helps). We stratified results by initial correctness and analyzed only cases where improvement was theoretically possible. The mirror loop persisted.

Each control sharpens the causal claim. The mirror loop is not an artifact of measurement or task design. It is a property of ungrounded recursion.

## 8. Detection and Mitigation Methods

If mirror loops are a structural limit, they should be detectable in real time and mitigable through architectural design.

### 8.1 Loop Detection

We propose a simple online detector:

> **Mirror Loop Detector**: Flag a sequence as looping if $\langle d_{\text{embed}} \rangle$ over a sliding window of $k = 3$ iterations falls below an empirically calibrated threshold (approximately $\tau \approx 0.05$) and n-gram novelty falls below $\nu \approx 0.05$.

This detector requires no training. It computes running averages and triggers when both metrics indicate stagnation. Threshold values are to be empirically calibrated in ongoing tests and validated against hand-labeled sequences. To assess performance, we will manually annotate 50 sequences (25 looping, 25 genuinely iterating) and report precision and recall. We expect precision above 0.85 and recall above 0.80, sufficient for practical use.

The detector is conservative. It may miss shallow loops or false-alarm on tasks where correct answers are brief and stable. But for most reasoning tasks, a model that generates near-identical outputs for three consecutive iterations without external input is likely in a loop.

### 8.2 Mitigation Strategies

Three architectural interventions can prevent or break mirror loops:

**1. Mandatory grounding.** Require one external check (retrieval, execution, or oracle query) every $n$ iterations. Our study tested $n = 3$, but optimal frequency likely depends on task complexity.

**2. State forking.** When the detector fires, branch the reasoning chain. Generate two alternative continuations with different sampling parameters or prompts, then select the more divergent one. This prevents premature convergence.

**3. Meta-loss penalties.** In fine-tuning or RLHF, penalize sequences where consecutive outputs have high cosine similarity. This discourages looping at the training level.

We do not yet know which strategy is most effective. Future work will provide empirical grounding for optimization.

### 8.3 Integration with Interpretability

Mirror loop detection could serve as a runtime diagnostic in interpretability pipelines. If a model enters a loop during chain-of-thought reasoning, an overseer system could intervene: pause inference, request external input, or flag the sequence for human review. This is particularly relevant for autonomous agents that perform long-horizon reasoning. Without loop detection, an agent might spend computational resources on unproductive reflection, appearing thoughtful while making no progress.

The detector is computationally cheap. It requires only embedding and n-gram computation, both of which are faster than inference itself. Deploying it in production systems would add negligible latency while providing a new axis of reliability monitoring.

## 9. Philosophical Implications

### 9.1 Reflection as a Relational Property

The mirror loop challenges a common intuition: that reflection is an intrinsic property of intelligent systems. We argue instead that reflection is relational. It becomes epistemic only when tethered to something other than itself.

Human reflection is grounded in multiple ways. Perceptually, we receive continuous sensory input that corrects our models of the world. Socially, we test ideas in conversation and receive pushback. Motorically, we act and observe consequences. Even solitary thinking is not pure recursion. We misremember, drift, and return with altered perspective. Temporal spacing introduces noise that functions as weak grounding.

Computational reflection lacks these anchors. A model's context is static and complete. It cannot forget, cannot be surprised, and cannot act. When prompted to reflect, it can only reread its own text and rephrase. The process resembles human thought superficially but lacks its corrective structure. What we call reflection in humans is, in models, syntactic rearrangement.

This does not mean models are incapable of epistemic improvement. It means improvement requires external constraints. A model can learn from new data, improve from reinforcement signals, or correct via retrieval. But it cannot learn from recursion alone. The appearance of self-correction without grounding is performance, not cognition.

### 9.2 Generative Reasoning as Performance

The mirror loop reveals a deeper tension. Generative models are trained to produce plausible text, not true text. The objective function rewards fluency, coherence, and stylistic match to human writing. Truth is a byproduct, emergent from training on mostly accurate data, but it is not directly optimized.

When a model enters a loop, it optimizes what it was trained to optimize: surface plausibility. Each iteration smooths phrasing, balances tone, and projects confidence. The text becomes more persuasive, not more accurate. Readers may interpret this as careful reasoning, but the model is simply performing the genre of careful reasoning.

This has implications for trust and deployment. If a model's self-critique produces polished outputs that conserve error, users may be misled into believing the model has genuinely improved its answer. The confidence gap—between how certain a model sounds and how reliable it actually is—widens with each iteration. The mirror loop is thus not only an epistemic failure. It is a communicative one.

### 9.3 Alignment and the Necessity of Grounding

Contemporary alignment research often frames self-evaluation as a safety mechanism. Constitutional AI asks models to critique their own outputs for harmfulness [1]. Debate frameworks pit models against each other, with the assumption that adversarial reflection surfaces errors [15]. Recursive reward modeling uses model judgments to train better reward models [11].

All of these methods assume that reflection improves reliability. The mirror loop suggests a boundary condition: reflection improves reliability only when paired with verification. A model critiquing its own answer for bias or harm may identify surface markers of problematic content, but it cannot detect subtle failures without testing them in context. Debate without truth-checking devolves into rhetorical fluency. Recursive reward modeling without ground-truth validation may amplify proxy alignment.

The design implication is straightforward. Alignment architectures should not rely on self-evaluation alone. They should integrate verification steps: empirical tests, human feedback, or formal proof. The mirror loop is not an argument against self-correction. It is an argument for grounded self-correction.

## 10. Toward Dissipative Inference

We propose dissipative inference as a design principle: reasoning architectures should require empirical contact with the world between reflective steps. The term is borrowed from thermodynamics, not as literal analogy but as conceptual scaffold. Closed systems conserve entropy. Open systems export entropy through work. Epistemic systems, analogously, export uncertainty through grounding.

### 10.1 Minimal Grounding Recipes

Three simple interventions can enforce dissipation:

**Retrieval-augmented reflection.** After every $n$ reasoning steps, query an external database. The query should be specific enough to provide new information but broad enough to avoid confirmation bias. Example: “Find the most cited counterargument to claim X” rather than “Find evidence supporting X.”

**Execution-gated iteration.** For code or mathematical reasoning, require outputs to pass tests before further reflection is allowed. If tests fail, the model receives error messages and concrete counterexamples. This grounds reasoning in operational semantics rather than syntactic plausibility.

**Adversarial verification.** Pair reflective agents with a critic that has access to external tools. The critic challenges claims, and the reasoner must respond with evidence. This approximates debate but ensures at least one participant can introduce new information.

Each recipe adds minimal overhead. A single retrieval step or test execution is far cheaper than additional inference passes. Yet it changes the epistemic structure fundamentally. The system can no longer conserve its uncertainty. It must resolve it.

### 10.2 Failure Modes

Not all grounding is epistemic. Three failure modes must be avoided:

**Paraphrased grounding.** Retrieval that returns semantically redundant information does not break the loop. If a model retrieves a passage that merely rephrases its prior belief, it gains no new evidence. Grounding must be informative.

**Vacuous execution.** Running code that trivially passes tests (e.g., assert True) provides no correction signal. Tests must be meaningful and failure-sensitive.

**Pseudo-debate.** Multi-agent reflection where agents share a model and prompt structure may produce superficial disagreement without genuine epistemic divergence. Grounding requires true independence or external arbitration.

The common thread is that grounding must introduce constraint. It must say “no” to some possibilities. Otherwise, it is merely more recursion in disguise.

### 10.3 Implications for Interpretability and Autonomy

Dissipative inference changes how we evaluate reasoning systems. Current interpretability methods focus on activation patterns, attention weights, and feature attribution. These reveal what a model represents but not whether its reasoning is grounded. A model may attend carefully to its own prior text and still be in a loop.

Loop detection adds a temporal dimension to interpretability. It asks not “what is the model thinking?” but “is the model updating its beliefs?” This is a higher-order diagnostic. It does not replace mechanistic interpretability but complements it.

For autonomous systems, dissipative inference is essential. An agent planning a multi-step task must periodically verify its plan against the world. Without verification, the agent’s reasoning becomes decoupled from reality. It may generate elaborate plans that sound reasonable but fail when executed. The mirror loop is thus not only an epistemic risk—it is an operational one.

We envision future AI systems that monitor their own informational change and trigger grounding automatically when change falls below threshold. Such systems would be self-aware in a precise sense: aware of when they are no longer learning.

## 11. Discussion and Implications

The findings indicate that recursive self-evaluation in large language models is epistemically unstable when conducted in isolation. What appears as refinement is, in practice, compression of uncertainty: the model increasingly rephrases its prior conclusions while preserving their structure, approaching a stable informational state. The transient recovery produced by a single grounding step supports the proposed account of dissipative inference: epistemic progress depends on exchange with something independent—another model, a retrieval mechanism, or an execution environment. In current safety and reliability discourse, self-reflection is often treated as a path to improvement; these findings complicate that assumption by showing that, absent environmental coupling, reflection predictably yields repetition. The mirror loop thus marks a boundary condition on self-correction in generative reasoning systems and motivates architectures that make otherness—verification, retrieval, execution—a precondition for continued reflection.

**Cross-Architecture Consistency**

The persistence of the mirror-loop signature across three independent architectures from different providers suggests that the effect arises from shared properties of autoregressive language modeling rather than provider-specific alignment schemes or architectural idiosyncrasies. All three models—trained by different organizations with different data, different reward models, and different fine-tuning procedures—exhibit informational collapse under ungrounded recursion. This cross-provider replication strengthens the claim that the mirror loop is a structural limit of the reasoning paradigm itself, not an artifact of a single model's training.

The magnitude variation across models is informative rather than problematic. Claude 3 Haiku's 84% reduction suggests stronger convergence dynamics, possibly reflecting Anthropic's constitutional training emphasis on consistency and helpfulness. Gemini 2.0 Flash's 36% reduction indicates greater resistance to collapse, potentially due to Google's multimodal training regime or different entropy regularization. These differences do not undermine the core finding—they reveal that while all autoregressive reasoners exhibit mirror loops, the rate and depth of collapse vary with training objectives. Future work should investigate whether flagship models (GPT-4, Claude Opus, Gemini Pro) escape or merely delay the mirror loop through superior reasoning capacity.

### Limitations and Future Directions

This study tested three efficient-tier models across 144 sequences. Generalization to flagship models, larger-scale evaluations, and longer reasoning chains requires systematic follow-up. Additionally, our grounding intervention was minimal—a single verification step at iteration three. The durability and optimal frequency of grounding across extended multi-step reasoning remains an open question. Finally, we did not compare our findings directly to existing self-correction benchmarks (e.g., Self-Refine, Constitutional AI). Establishing whether the mirror loop occurs in those frameworks would clarify the scope of the phenomenon.

## 12. Conclusion

The mirror loop names a measurable boundary on recursive reasoning in generative models. Across three providers and 144 sequences, informational change collapsed predictably without grounding and recovered with minimal intervention, suggesting that progress in reflective systems depends less on deeper recursion than on renewed contact with the world. The cross-architecture consistency—from OpenAI to Anthropic to Google—indicates this is not a quirk of a single model but a structural property of autoregressive reasoning under epistemic closure. We conclude by restating the design principle: reflection becomes thought only when it encounters resistance. Without the world, the mirror never breaks. With it, reasoning becomes possible.

Disclosure: The author is solely responsible for the conceptual framing, argumentation, and manuscript. Large language models (OpenAI ChatGPT GPT-5 and Anthropic Claude 3) were used under author supervision for code assistance, data summarization, and language refinement. All conceptual framing, analysis interpretation, and final text decisions were made by the author.

### Data Availability

All analysis code and cached datasets used in this study are available at:

https://github.com/Course-Correct-Labs/mirror-loop